\documentclass[a4paper,10pt]{article}
\usepackage[utf8]{inputenc}
\pdfoutput=1
\usepackage{hyperref}
\hypersetup{pdfauthor={Tamas Hajgato},pdftitle={review of compressed embedding layers and their applications for recommender systems}}
\usepackage{float}
\usepackage{graphicx}
\usepackage{amssymb}
\usepackage[margin=1.5in]{geometry}

\DeclareFixedFont{\ttb}{T1}{txtt}{bx}{n}{9} % for bold
\DeclareFixedFont{\ttm}{T1}{txtt}{m}{n}{9}  % for normal

\usepackage{listings}

\title{Review of compressed embedding layers and their applications for recommender systems}
\author{Tamas Hajgato}

\begin{document}

\maketitle

\begin{abstract}
We review the literature on trainable, compressed embedding layers and discuss their applicability for compressing gigantic neural recommender systems. We also report the results we measured with our compressed embedding layers. 
\end{abstract}

\section{Introduction} \label{sec-introduction}

Information technology has become widely spread in industrial applications. Extraordinarily large amounts of data have been made accessible to users. This has made it difficult to select the data that the user needs. One possible resolution of this issue came from the field of deep learning, from the discovery of recommender systems. \cite{recommenders} These systems help users go through the process of decision-making and selecting the relevant data. 

A recommender system predicts the users behavior in order to detect their interests and needs, and the relevant data can then be recommended. The state of the art recommender systems typically incorporate embedding layers with enormous sizes. This appears to limit the choice of hardware to run these systems. Embedding layers may not fit into GPU memory. There are various, more-or-less straightforward alternative solutions: 

\begin{itemize}
 \item 
One such solution is to store the embedding layers outside of GPU memory, and to apply an efficient caching mechanism in an attempt to minimize the performance hit. \cite{microsoft2}, \cite{parameterserver}, \cite{caching}
 \item 
An other solution is to use multiple GPUs and distribute the huge embedding layers between them. This can be expected to be more efficient, as the communication between multiple GPUs can be done more efficiently than copying large chunks of data from CPU to GPU memory whenever a cache-miss occurs. Storing embeddings on a parameter server can be expected to be even less efficient than storing them in CPU memory. \cite{nvidia}
 \item 
Of course, alternative hardware could also be used. However, due to the enormous size of the embedding layers and the heavy-tailed access pattern of features, it's reasonable to expect that inference will be bottlenecked by memory operations. 
\end{itemize}

In this paper we attempt to address this problem by introducing a trainable, compressed embedding layer. Intuitively, our solution transforms the enormous size (and the cost of memory operations that come with it) into additional tensor contractions. We show that it is possible to replace enormous embedding layers with significantly smaller, so called \emph{compressed}, or rather \emph{decomposed embedding layers} without changing the hyperparameters of the network. This replacement comes without any accuracy losses, at the expense of a negligible latency hit. This, in turn it gives us huge throughput gains when using modern GPUs, since recommenders typically leave efficient GPUs starving for compute tasks. See \cite{performance}. Moreover, it gives us significant savings in both hardware and operational costs of running recommender systems. 

It must be noted, that our method can be used in conjunction with the parameter servers and efficient caching methods already available. See \cite{caching}. 

\section{Review of previous work}

Let's review the previous \emph{algorithmic} attempts at solving this problem. See the papers \cite{facebook}, \cite{saec}, \cite{facebook2}, \cite{microsoft} and \cite{PSS}. \\

The authors of \cite{facebook} compare various int4 quantization approaches in Table 2 of \cite{facebook}.  They use normalized L2 loss for the metric of comparison. The best methods (with respect to this metric) require clusterization of a large amount of data. Moreover, the paper claims that the accuracy of the compressed model is the same as the original one - however, in the paper only the values of the loss function were compared. We suspect, that in some cases the network may need additional training in order to offset the accuracy loss that comes from the int4 quantization of the embeddings. Clustering alone requires a lot of compute on top of training the network, in exchange for a theoretical maximum of 8x size reduction (when going from the 32-bit floating point format to the 4-bit integer format), and a smaller expected size reduction. The method we used is a better alternative. \\

The algorithm in \cite{saec} is a version of k-means clustering, that relies on the empirical heavy-tailed nature of features. The initial values of the centroids for k-means are the most frequent rows of the embedding. As a consequence, the same vector may represent multiple embedding vectors. There is no quantization in \cite{saec}. The claims of this paper are remarkable, however it must  be noted that the algorithm was intended to be used when limitless data is available. For this reason, it's difficult to compare the accuracy of the compressed model to the original one - since in this paper they were trained on different datasets. We suspect that compression harms the accuracy curve and we would be able to observe this only if we compared the original model to the compressed one, and both models were trained on the exact same dataset, and until the accuracy can improve. Under these conditions we will likely find that the compressed model performs worse than the original. We claim that this can be reasonably expected, since the algorithm here creates a few representative embedding vectors to use in place of the original embedding vectors, and attempts to offset the accuracy loss with further training. Therefore, the implicit assumption here is that this can always be done without any accuracy loss. The kind of measurement that would properly determine whether this algorithm comes with an accuracy loss was omitted from \cite{saec}. \\

We found the idea to store the embedding weights in a tensor-train format in \cite{facebook2}. The weights are not decomposed after training, but are kept in tensor-train format throughout training and inference. Therefore, the network has to be able to learn the decomposition. Moreover, the requested entries of the weight matrix have to be computed at inference time. This computation consists of a few lookups followed by tensor contractions. We are going to refer to this idea and its derivatives as \emph{decomposed training}. 
The authors of \cite{facebook2} implemented highly optimized kernels and a caching mechanism in an attempt to speed up the lookup procedure. Moreover, they described (an approximation of) a few popular initialization methods. The intention here is to use these approximations on the tensor-train as substitutes for the initialization methods often used on embedding weights. 

From the bar-plots published in \cite{facebook2} it is unclear whether the application of this method comes with an accuracy loss. Moreover, it is unclear what kind of accuracy metric is used in the paper. Page 7, right before Chapter 6.1 of \cite{facebook2} states the following. \emph{``Finally, the model quality experiences negligible degradation. For example, for both Kaggle and Terabyte, using TT-Rec to train the five largest embedding tables in the TTEmb. format using TT-rank of 32 leads to model accuracy loss within 0.03\%. ''} Both the size of the tensor-train embedding and the extent of the accuracy degradation depend on the hyperparameters of the tensor-train algorithm. \cite{facebook2} does not discuss any kind of a heuristic for choosing the hyperparameters. It is therefore left to the user to figure this out. There is also the question of whether the tensor-train algorithm is an appropriate choice for this kind of an approach, and how exactly does it compare to other, similar methods. 

For most industrial applications some negligible accuracy loss is acceptable. This is not the case for recommenders. Therefore, according to our judgement \cite{facebook2} is an initial investigation. It leaves a lot of questions open, and doesn't offer a solution that is applicable in the industry. Perhaps the same can be said about an earlier work, found in \cite{tensorized}, which introduced TT-embeddings for MLPs, LSTMs and transformers. The initialization procedure in \cite{tensorized} seems to be the similar to the one found in \cite{facebook2}, and so it suffers from the same faults. 

Our initialization procedure yields the kind of randomness the user expects. A lot of thought went into making our technology future-proof. We offer a solution that can be expected to work well with other neural networks, and to scale way beyond the neural network sizes we tested it on. \\

The algorithm found in \cite{microsoft} is another attempt at inventing decomposed embeddings. The paper introduced the so called ``MEmCom algorithm (with no bias)'' as (experimentally) an improvement over the ``quotient-remainder trick'' of \cite{shi}. Both methods fall into the category of low-rank decompositions of tensors. We have carefully reviewed the algorithms proposed by \cite{microsoft} and \cite{shi}, and found that both of them came with worse accuracy results than our method. The mathematical intuition behind the ``MEmCom algorithm'' was not discussed in \cite{microsoft}. These algorithms are significantly simpler than the one in \cite{facebook2} and come with no new hyperparameters. They might work --- however, after a careful mathematical analysis we have found their applicability too limiting perhaps. \\

We've found a number of other attempts at inventing decomposed embeddings. See \cite{nlp} for more references. These attempts were intended to be used in different areas of deep learning, for example, for word representations in NLP models. They all seem to come with various degrees of accuracy losses and poor initialization procedures. Therefore, they are not meant to be used for recommenders, where even a slight loss in accuracy results in a measurable loss of revenue. Moreover, the initialization of the embeddings is important for recommenders. \cite{init} \\

The work published in \cite{wang} explored the idea of storing weights in a tensor-ring format, instead of a tensor train. However, \cite{wang} investigated the accuracy trade-off for fully-connected and convolution layers only. The conclusion in \cite{wang} was that tensor-ring can preserve the weights with moderately lower compression ratios than that of tensor-train, while being computationally more expensive. \\

The paper \cite{PSS} proposed the idea to map the embedding vectors onto a low-dimensional manifold with the use of the well-known Johnson-Lindenstrauss Lemma. Due to this lemma the pairwise distances between the embedding vectors are (more-or-less) preserved. At inference time the embedding vectors are mapped back into the original dimension with the transpose of the Johnson-Lindenstrauss transform. This idea is improved by segmenting the embedding vectors into chunks and approximating the chunks independently. 

The main assumption here is that if we preserve the pairwise distances between the embedding vectors, then we will also preserve the accuracy of the neural network. We believe that this is not true. There is no reason for the accuracy of the neural network to depend only on the pairwise distances between the embedding vectors. In general, there is no reason for any matrix sketching algorithm to preserve the accuracy of the network. It might work though. 
It must be noted, that inference is (more-or-less) a single tensor contraction --- just like in our method. Moreover, the compression ratio we have seems to be similar. Therefore, we don't expect huge latency and throughput differences. As for the convergence of the training, our method requires more compute power, but it also comes with more guarantees. As a consequence, we train for less number of epochs. \cite{PSS} reported training DLRM for 12 epochs, as opposed to our 2-3 epochs. Due to the simplicity of their backward, the overall time they needed to train was less than what we needed. However, we expect that when the size of the network grows even larger, then even the compressed network will be large. As a consequence, it will be more difficult to utilize the hardware, and so training for 12 epochs will not be a quick task anymore. 

It must be noted, that the claim of \cite{PSS} is that this algorithm is able to compress the DLRM model 10000$\times$ without any accuracy losses. However, we can see an accuracy drop in the tables in Figure 4 of \cite{PSS}. It must also be noted, that for industrial applications the compression ratio might be less important --- accuracy and throughput (subject to an upper bound on latency) are more important. The paper does not discuss any kind of initialization method, nor any heuristics for selecting the new hyperparameters. \\

Our work corrects all the shortcomings of the other works. We took the ideas published in \cite{facebook2}, and after a careful mathematical analysis we somehow managed to develop a decomposed embedding layer that satisfies all the requirements for industrial application. We replace the native embedding layers of PyTorch with our decomposed embedding layers. Under reasonable conditions this replacement will not need any hyperparameters to be changed, and the accuracy of the network will be preserved. We've worked out the appropriate mathematical discussion relevant to that. Unfortunately, our decomposed embedding layers depend on some new hyperparameters. 

We made a simple heuristic for pre-selecting the hyperparameter configurations that are likely to yield good results under reasonable conditions. We theorize, that on real-world datasets it may be possible to use our decomposed layers under the hood - and let our (or a similar) simple formula set the hyperparmeters. Verifying such a hypothesis however is beyond the scope of what we wanted to accomplish. 

\section{Our results} \label{sec-results}

In this chapter we present our experimental results. We have replaced some of the embedding layers in both NCF and DLRM in NVIDIA's open sourced DeepLearningExamples repository. NCF was trained on the MovieLens-1M, while DLRM was trained on the Criteo-1TB dataset. Here we only present our results for DLRM. Training was done with batch size 65536. Latency, throughput and accuracy measurements were done after we converted the trained model into TensorRT format for deployment. 

\subsection{Setting the hyperparameters}

We have introduced the {\ttm Frobenius\_layer} with the intent to serve as a drop-in replacement of the {\ttm native} embedding layers. We made efforts to future-proof our technology. It is natural to ask, if we succeeded. In order to say something about that, we have replaced the \textbf{four largest} embedding layers in DLRM with the {\ttm Frobenius\_layer}. Below we present our measurements with various hyperparameter configurations and a fixed random seed. 
\begin{table}[H]
\centering
 \begin{tabular}{||c c c c c c||}
 \hline
 layer & {\ttm r} & {\ttm p} & size (mb) & epochs & accuracy (auc) \\
 \hline\hline
 {\ttm native} & --- & --- & 14057.3007 & 2 & 0.8025 \\
 \hline
 {\ttm F} & 8  & 1 & 0.6699 & 2 & 0.8032 \\
 {\ttm F} & 8  & 2 & 1.3243 & 2 & 0.8034 \\
 {\ttm F} & 8  & 4 & 2.6328 & 2 & 0.8037 \\
 {\ttm F} & 8  & 8 & 5.2490 & 2 & 0.8033 \\
 \hline
 {\ttm F} & 16 & 1 & 1.3396 & 2 & 0.8032 \\
 {\ttm F} & 16 & 2 & 2.6484 & 2 & 0.8034 \\
 {\ttm F} & 16 & 4 & 5.2650 & 2 & 0.8036 \\
 {\ttm F} & 16 & 8 & 10.500 & 2 & 0.8033 \\
 \hline
 \end{tabular}
\end{table}

In the table above {\ttm F} stands for {\ttm Frobenius\_layer}. Moreover, ``size'' refers to the total size of the four embedding layers before/after replacement, respectively. The parameters of all of the embeddings are stored in 32-bit floating-point format. The total size of the rest of the embeddings is 1913.9582 mb. Below is a comparison of the accuracy (auc) curves. 

\begin{figure}[H]
\centering
\includegraphics[height=0.53\textwidth,width=1.0\textwidth]{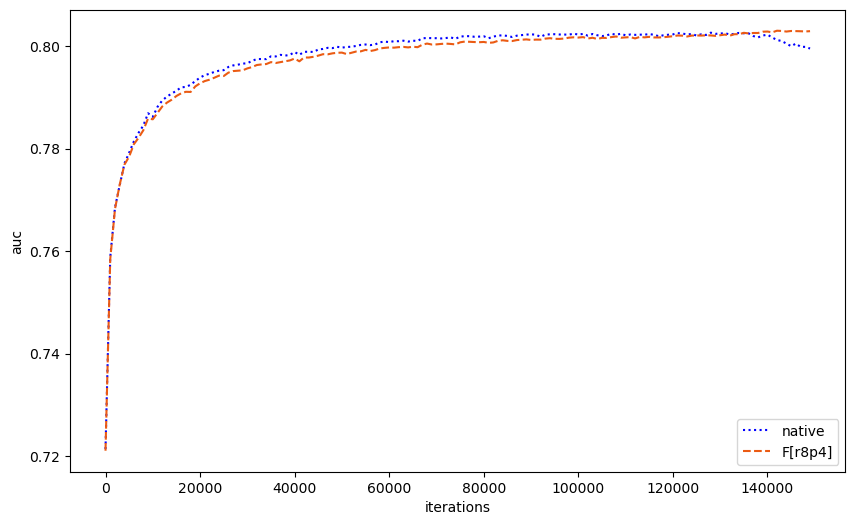}
\end{figure}

Observe the drop on the accuracy graph of DLRM above. The following is happening here. After some iterations the validation loss starts to increase, while the training loss keeps decreasing - that is, DLRM is overfitting the data. We theorize, that perhaps {\ttm Frobenius\_layers} function as regularizers. \\

It must be noted here, that the authors of \cite{facebook2} made use of a smart caching mechanism in their solution. Thanks to this idea, the training procedure was able to optimize the few embedding vectors in the cache directly - and perhaps independently from the embedding layers. Perhaps, such a mechanism can be expected to speed up the training process - and it indeed might be a good idea, when the embedding layers are small or the access pattern of the features is not heavy-tailed. However, such ideas tend to ignore the mathematics of the training problem. We argue, that perhaps choosing the better accuracy in exchange for the longer training time is worth it. \\
\quad \\
It is natural to ask, what would happen if we replaced \textbf{five} instead of the four largest embedding layers. We present our measurements below. 
\begin{table}[H]
\centering
 \begin{tabular}{||c c c c c c||}
 \hline
 layer & {\ttm r} & {\ttm p} & size (mb) & epochs & accuracy (auc) \\
 \hline\hline
 {\ttm native} & --- & --- & 15363.912 & 2 & 0.8025 \\
 \hline
 {\ttm F} & 8 & 1 & 0.7739 & 3 & 0.8029 \\
 {\ttm F} & 8 & 2 & 1.5284 & 3 & 0.8029 \\
 {\ttm F} & 8 & 4 & 3.0371 & 3 & 0.8032 \\
 {\ttm F} & 8 & 8 & 6.0537 & 3 & 0.8029 \\
 \hline
 {\ttm F} & 16 & 1 & 1.5476 & 3 & 0.8028 \\
 {\ttm F} & 16 & 2 & 3.0566 & 3 & 0.8029 \\
 {\ttm F} & 16 & 4 & 6.0736 & 3 & 0.8031 \\
 {\ttm F} & 16 & 8 & 12.119 & 3 & 0.8029 \\
 \hline
 \end{tabular}
\end{table}
Here, the total size of the rest of the embeddings is 607.3469 mb. It must be noted, that this time it was necessary to train for 3 epochs to match the native accuracy. \\
\quad \\
Based on the measurements above, perhaps for similar recommender sizes both {\ttm r} and {\ttm p} can be set to small integers. Moreover, it seems that the {\ttm Frobenius\_layer} is functioning as intended. In theory, {\ttm r} can be large. It seems that instead of increasing {\ttm r} to a potentially large value, we can increase {\ttm p} in order to improve the capacity of the model. For each of the replaced embeddings above, our heuristic recommended the following {\ttm (r, p)} pairs: (24, 1), (32, 1), (16, 2) and (8, 4). Out of these, our algorithm selected (8, 4) by default. In the rest of the paper we are going to let the heuristic select the hyperparameters for us. 

\subsection{Training}

\begin{table}[H]
\centering
 \begin{tabular}{||c c c c c c||}
 \hline
 variant & GPUs & layer & epochs & accuracy (auc) & training time \\
 \hline\hline
 small & 1xA100(40GB) & {\ttm native} & 2 & 0.8034 & 57min \\
 large & 1xA100(40GB) & {\ttm native} & --- & --- & --- \\
 xlarge & 1xA100(40GB) & {\ttm native} & --- & --- & --- \\
 full & 1xA100(40GB) & {\ttm native} & --- & --- & --- \\
 \hline
 small & 1xA100(40GB) & {\ttm F[4]} & 2 & 0.8036 & 66min \\
 large & 1xA100(40GB) & {\ttm F[5]} & 3 & 0.8030 & 100min \\
 xlarge & 1xA100(40GB) & {\ttm F[5]} & 3 & 0.8029 & 102min \\
 full & 1xA100(40GB) & {\ttm F[5]} & 3 & 0.8028 & 101min \\
 \hline
\end{tabular}
\end{table}

\begin{table}[H]
\centering
 \begin{tabular}{||c c c c c c||}
 \hline
 variant & GPUs & layer & epochs & accuracy (auc) & training time \\
 \hline\hline
 small & 1xH100(80GB) & {\ttm native} & 2 & 0.8025 & 58min \\
 large & 1xH100(80GB) & {\ttm native} & --- & --- & --- \\
 xlarge & 1xH100(80GB) & {\ttm native} & --- & --- & --- \\
 full & 1xH100(80GB) & {\ttm native} & --- & --- & --- \\
 \hline
 small & 1xH100(80GB) & {\ttm F[4]} & 2 & 0.8032 & 65min \\
 large & 1xH100(80GB) & {\ttm F[5]} & 3 & 0.8027 & 101min \\
 xlarge & 1xH100(80GB) & {\ttm F[5]} & 3 & 0.8029 & 101min \\
 full & 1xH100(80GB) & {\ttm F[5]} & 3 & 0.8030 & 100min \\
 \hline
 \end{tabular}
\end{table}
In the tables above {\ttm F[5]} means that the 5 largest embedding layers were replaced with {\ttm Frobenius\_layers}, respectively. Moreover, the ``variant'' setting above refers to the frequency thresholding of the dataset, as described in NVIDIA's open source DeepLearningExamples repository, see \cite{deeplearningexamples-dlrm}. For the sake of completeness we have included the relevant table here: 
\begin{table}[H]
\centering
 \begin{tabular}{||c c c||}
 \hline
 variant & frequency threshold & uncompressed model size \\
 \hline\hline
 small & 15 & 15GB \\
 large & 3 & 82GB \\
 xlarge & 2 & 142GB \\
 full & 0 & 421GB \\
 \hline
 \end{tabular}
\end{table}
It's important to know if the results presented in the table above are easy to reproduce. To say something about this we trained the model with different random seeds. \begin{table}[H]
\centering
 \begin{tabular}{||c c c c c c c||}
 \hline
 variant & layer & auc(seed1) & auc(seed2) & auc(seed3) & auc(seed4) & auc(seed5) \\
 \hline\hline
 small & {\ttm native} & 0.8025 & 0.8036 & 0.8034 & 0.8035 & 0.8029 \\
 \hline\hline
 small & {\ttm F[4]} & 0.8032 & 0.8030 & 0.8036 & 0.8032 & 0.8034 \\
 \hline
 large & {\ttm F[5]} & 0.8027 & 0.8032 & 0.8030 & 0.8032 & 0.8029 \\
 \hline
 xlarge & {\ttm F[5]} & 0.8029 & 0.8031 & 0.8029 & 0.8030 & 0.8029 \\
 \hline
 full & {\ttm F[5]} & 0.8030 & 0.8026 & 0.8028 & 0.8029 & 0.8030 \\
 \hline
 \end{tabular}
\end{table}
The following table compares the sizes of the compressed models above to the sizes of the corresponding uncompressed models, respectively. 
\begin{table}[H]
\centering
 \begin{tabular}{||c c c||}
 \hline
 variant & uncompressed model size & compressed model size \\
 \hline\hline
 small & 15GB & 0.93GB \\
 large & 82GB & 0.58GB \\
 xlarge & 142GB & 0.70GB \\
 full & 421GB & 0.99GB \\
 \hline
 \end{tabular}
\end{table}

\subsection{Performance}

The graph below shows which batch size saturates the GPU with respect to the various frequency filtering settings and layer combinations. It must be noted, that latency, throughput and accuracy measurements were done after we converted the trained model into TensorRT format. The numbers above the dots show the batch size that belongs to the corresponding throughput/latency values. 

\begin{figure}[H]
\centering
\includegraphics[height=0.53\textwidth,width=1.0\textwidth]{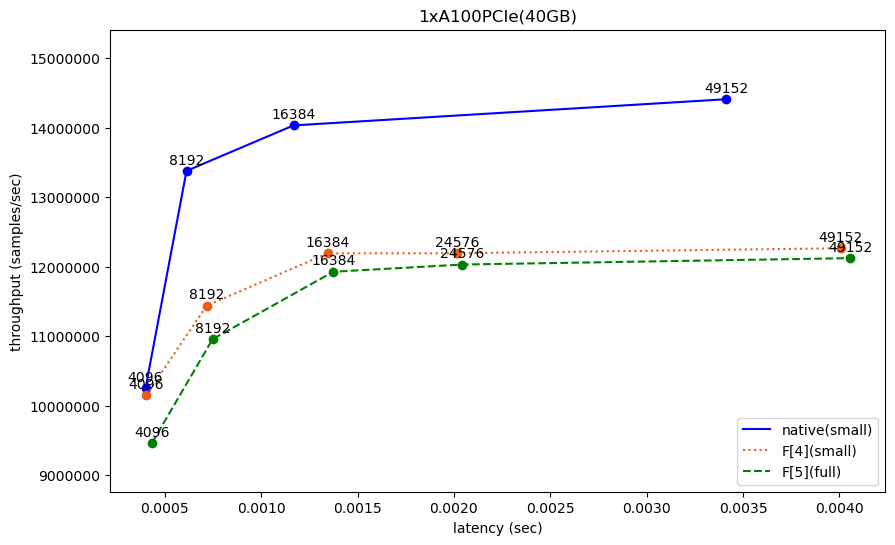}
\end{figure}

We can see on the graph above that the GPU is already saturated with {\bf batch size 49152}. A batch size that's double that can be expected to take twice as long to process. In the table below we show the numeric values that correspond to batch size 49152. 

\begin{table}[H]
\centering
 \begin{tabular}{||c c c c c c||}
 \hline
 variant & GPUs & layer & auc & latency & throughput \\ 
 \hline\hline
 small & 1xA100(40GB) & {\ttm native} & 0.8034 & 0.00341179s & 14,406,513 samples/s \\
 large & 1xA100(40GB) & {\ttm native} & --- & --- & --- \\
 xlarge & 1xA100(40GB) & {\ttm native} & --- & --- & --- \\
 full & 1xA100(40GB) & {\ttm native} & --- & --- & --- \\
 \hline
 small & 1xA100(40GB) & {\ttm F[4]} & 0.8036 & 0.00400808s & 12,263,228 samples/s \\
 large & 1xA100(40GB) & {\ttm F[5]} & 0.8030 & 0.00416105s & 11,812,403 samples/s \\
 xlarge & 1xA100(40GB) & {\ttm F[5]} & 0.8029 & 0.00407425s & 12,064,060 samples/s \\
 full & 1xA100(40GB) & {\ttm F[5]} & 0.8028 & 0.00405516s & 12,120,853 samples/s \\
 \hline\hline
 small & 1xH100(80GB) & {\ttm native} & 0.8025 & 0.00219485s & 22,394,241 samples/s \\
 large & 1xH100(80GB) & {\ttm native} & --- & --- & --- \\
 xlarge & 1xH100(80GB) & {\ttm native} & --- & --- & --- \\
 full & 1xH100(80GB) & {\ttm native} & --- & --- & --- \\
 \hline
 small & 1xH100(80GB) & {\ttm F[4]} & 0.8032 & 0.00259026s & 18,975,701 samples/s \\
 large & 1xH100(80GB) & {\ttm F[5]} & 0.8027 & 0.00267112s & 18,401,269 samples/s \\
 xlarge & 1xH100(80GB) & {\ttm F[5]} & 0.8029 & 0.00265543s & 18,509,996 samples/s \\
 full & 1xH100(80GB) & {\ttm F[5]} & 0.8030 & 0.00265716s & 18,497,945 samples/s \\
 \hline
 \end{tabular}
\end{table}

Keep in mind, that everything presented here (including accuracy) was measured after the model was deployed into TensorRT format. \\

For the sake of making an ``apple-to-apple'' comparison, let's include the results published by NVIDIA for the MLPerf competition Round v3.0 (2023), see \cite{mlperf2023}. The performance numbers in the table below were measured on an 8xH100 PCIe-80GB setup. Thus, we were able to use 8 instances of the compressed DLRM model. 

\begin{table}[H]
\centering
 \begin{tabular}{||c c c c c||}
 \hline
 variant & GPUs & layer & auc & throughput \\ 
 \hline\hline
 ? & 8xH100(80GB) & {\ttm native} & 0.8025 & 3,713,980 samples/s \\
 \hline
 small & 8xH100(80GB) & {\ttm F} & 0.8032 & 151,805,608 samples/s \\
 large & 8xH100(80GB) & {\ttm F} & 0.8027 & 147,210,152 samples/s \\
 xlarge & 8xH100(80GB) & {\ttm F} & 0.8029 & 148,079,968 samples/s \\
 full & 8xH100(80GB) & {\ttm F} & 0.8030 & 147,983,560 samples/s \\
 \hline
 \end{tabular}
\end{table}

For the sake of completion let us mention here, that according to \cite{mlperf2023}, on an 8xH100 SXM-80GB setup the throughput of 5,366,820 samples/s was measured. It also has to be noted, that - as opposed to the submission to the MLPerf competition - we only made minimal changes to the DLRM implementation found in NVIDIA's DeepLearningExamples repository. \cite{deeplearningexamples-dlrm}

\end{document}